\pdfoutput=1

\documentclass[11pt]{article}

\usepackage[final]{acl}

\usepackage{times}
\usepackage{latexsym}

\usepackage[T1]{fontenc}

\usepackage[utf8]{inputenc}

\usepackage{microtype}

\usepackage{inconsolata}

\usepackage{graphicx}   
\usepackage{multirow}
\usepackage{multicol}
\usepackage{float}

\title{Detection and Positive Reconstruction of Cognitive Distortion sentences:\\Mandarin Dataset and Evaluation}

\author{Shuya Lin$^1$, Yuxiong Wang$^1$, Jonathan Dong$^2$, \and Shiguang Ni$^1$\thanks{Corresponding author: ni.shiguang@sz.tsinghua.edu.cn}  \\
        $^1$Shenzhen International Graduate School, Tsinghua University \\ $^2$École polytechnique fédérale de Lausanne\\}

\begin{document}
\maketitle

\begin{abstract}
This research introduces a Positive Reconstruction Framework based on positive psychology theory. Overcoming negative thoughts can be challenging, our objective is to address and reframe them through a positive reinterpretation. To tackle this challenge, a two-fold approach is necessary: identifying cognitive distortions and suggesting a positively reframed alternative while preserving the original thought's meaning. Recent studies have investigated the application of Natural Language Processing (NLP) models in English for each stage of this process. In this study, we emphasize the theoretical foundation for the Positive Reconstruction Framework, grounded in broaden-and-build theory. We provide a shared corpus containing 4001 instances for detecting cognitive distortions and 1900 instances for positive reconstruction in Mandarin. Leveraging recent NLP techniques, including transfer learning, fine-tuning pretrained networks, and prompt engineering, we demonstrate the effectiveness of automated tools for both tasks. In summary, our study contributes to multilingual positive reconstruction, highlighting the effectiveness of NLP in cognitive distortion detection and positive reconstruction. 
\end{abstract}

\section{Introduction}

The rapid pace of modern life and the prevalence of high-pressure lifestyles have led to a rise in mental health issues among individuals. According to a report by the World Health Organization (WHO), depression stands out as a significant contributor to both physical and mental disorders globally \cite{WHODepression2021}. Moreover, in response to the escalating concern, the WHO recently launched a new Commission on Social Connection in November 2023, recognizing the urgency of combating loneliness as a critical health hazard \cite{WHOloneliness2023}.
These indicators strongly suggest that within an environment marked by increased instability and diminished personal agency, there is a heightened risk for individuals to experience negative emotions, closely linked with negative thought patterns. These patterns, identified as cognitive distortions \cite{beck1963thinking}, represent rigid and inaccurate ways of thinking. Such detrimental thoughts reinforce self-negation and perpetuate negative emotions \cite{rnic2016cognitive}. These distorted thoughts, often irrational and reinforced over time, become deeply ingrained. Habitual adoption of such patterns makes them challenging to recognize. Consequently, they are profoundly destructive, when individuals perceive these distortions as truths. \par
Cognitive restructuring is the main method to work on cognitive distortions in a clinical setting \cite{dawes1964cognitive}. The process can be broadly categorized in two key phases: First, recognizing negative thinking patterns and mitigating their impact, this process requires individuals to write down their negative thoughts and identify any underlying cognitive distortions in those thoughts. Then, the cognitive restructuring process is designed to help cultivate a more objective view of the situation.\par
Yet, self-examining and rectifying distorted thinking can prove challenging for individuals \cite{beck1979cognitive}. Often, intervention and guidance by trained counselors are necessary to identify cognitive distortions and erroneous beliefs. Professional counselors can then follow with cognitive restructuring, drawing from their expertise to aid individuals in fostering new perspectives. This approach significantly relies on the counselor's unique skills and individual style. However, due to the high price of psychological counseling \cite{MHS2020} and the reluctance to speak with a therapist, most struggling people find it challenging to get the support they need. Even if they can afford the high cost of a consultation, they may have to wait months for an appointment \cite{mulraney2021long}. Therefore, in recent years, research has begun to turn its attention to Natural Language Processing (NLP) methods to assist individuals in reconstructing negative thoughts.\par
In recent years, the integration of NLP into the field of psychology has been gaining momentum. These NLP models, initially trained on large datasets for general purposes, are now being fine-tuned to address specific needs in mental health \cite{sharma2021computational,zheng2021comae,zhu2022multi,chen2022emphi,maddela2023training}. The emerging success of NLP in mental health is exemplified by applications \cite{prochaska2021therapeutic,chiauzzi2023relational,lin2023empathy,sharma2023cognitive}, demonstrating its potential for widespread, accessible psychological support. This development represents a significant advancement, offering a new avenue for cognitive restructuring and mental wellness, seamlessly connecting the high demand for mental health support with innovative, technology-driven solutions.\par
In this rapidly-evolving landscape, the scarcity of language-specific datasets poses a significant challenge. Currently, available datasets in the field of mental health NLP predominantly focus on the English language. Recognizing the linguistic and cultural intricacies inherent in mental health expressions, it becomes evident that dedicated datasets in other languages is essential for evaluating and fine-tuning NLP models. \par
In this study, we build and openly share the first Mandarin dataset\footnote{https://github.com/405200144/Dataset-of-Cognitive-Distortion-detection-and-Positive-Reconstruction/tree/main\label{web}} for both phases of the cognitive restructuring process, encompassing both cognitive distortion detection and positive reframing. Additionally, we benchmark different NLP models and training strategies for each task. Our main contributions are:
\begin{itemize}
  \item Construction of a Chinese dataset for the detection of cognitive distortions, based on real psychology Q\&A sentences and annotated by trained specialists.
  \item Construction of a Chinese dataset for positive reconstruction of cognitive-distorted sentences. Each sentence is reframed with five different strategies from the theory of positive psychology. 
  \item Benchmark of cognitive distortion detection using pre-trained RoBERTa-wwm-ext network with fine-tuning or transfer learning with different readout strategies (linear, multilayer perceptron, LSTM), with the best results obtained with fine-tuning the whole pre-trained network. 
  \item Benchmark of different approaches for positive reconstruction (P-Tuning, fine-tuning, prompt engineering) with both algorithmic and human evaluation, with the best results consistently across all metrics obtained with P-Tuning. We show also that among the different strategies possible in positive psychology, the best performing network predominantly chooses an optimistic strategy. 
\end{itemize}

\section{Related Works}

Previous studies have proposed various approaches to build NLP pipelines on the detection, categorization of distortions, and positive reframing. We detail here the challenges regarding corpus annotation, in particular labeling consistency, and the need for multilingual research. 

\subsection{Cognitive Distortion Detection}

The seminal classification of cognitive distortions \cite{beck1963thinking} has been revisited by recent studies. For instance, a study \cite{shickel2020automatic} used 15 categories from a psychological website to attempt classification. Similarly, \citet{shreevastava2021detecting} used a standard of 10 categories for classification, highlighting the complexity of corpus annotation for detection and categorization.  Another study \cite{tauscher2023automated} conducted a classification task with 5 categories, utilizing over 7,000 text messages in a study involving individuals with serious mental illness and clinicians.\par
However, these studies have revealed certain limitations. The study by \citet{shreevastava2021detecting}, for instance, reported only a 61\% internal consistency in detecting distortions. \citet{shickel2020automatic} encountered challenges in accurately classifying non-distorted cases. Furthermore, research of \citet{tauscher2023automated} had a low performance in the F1 score of merely 0.62. Importantly, a recent study \cite{maddela2023training} showcased the effectiveness of fine-tuned language model for cognitive distortion detection, achieving scores above 0.9 thanks to a large crowdsourced dataset. 
Similarly, Wang et al. \cite{wang2023c2d2} released a Chinese dataset in 2023, achieving a 0.73 F1 score, also leveraging a large crowdsourced dataset. However, a limitation of both studies is that the cognitive distortion sentences used were artificially created for the specific task rather than being derived from genuine conversations, potentially limiting the models' real-world applicability. 
\par
To sum up, recent research findings underscore the ongoing collective effort on the detection of cognitive distortions using language models. This fast-evolving landscape highlights the potential of large language models on this particular task, with a particular emphasis on the construction of a specialized dataset. 
We emphasize the importance of the detection task over classification due to its higher practical applicability. In a conversation, distortion detection may then activate the rest of the cognitive restructuring framework. Notably, there is currently no pertinent dataset available in the Chinese context, based on real-world dialogues, annotated by trained specialists. 


\subsection{Text Reframing}
Earlier works in text style transfer primarily focuses on altering semantic meaning, encompassing tasks such as rewriting biased articles \cite{ma2020powertransformer,reid2020automatically}, modifying proper nouns \cite{van2019evaluating}, and style transfer on comments from social media \cite{sudhakar2019transforming}. 

A study \cite{ziems2022inducing}, introduced the notion of employing text reconstruction to foster a positive outlook, utilizing stress-tagged tweets from Twitter. However, their reconstructing strategies were extracted from previous research \cite{harris2007integrating} focused more on forgiveness and spirituality in counselling practice. 
Instead, we believe that applying principles from the broaden-and-build theory can bolster the theoretical foundation of current and future works on the positive reframing task. 
Moreover, their approach uniformly rewrites all collected data, overlooking that all sentences do not necessarily present cognitive distortion.
Their strategy was used by another study \cite{maddela2023training}, which created around thirty thousand English sentences with unhelpful thought patterns and their rewritten counterpart. Nevertheless, they focus more on creating and reconstructing cognitive distortion texts, not detecting them from existing real-life data. 

Furthermore, \citet{sharma2023cognitive} introduced seven additional reconstruction strategies and curated a dataset comprising 1,077 samples labeled with thirteen cognitive distortion categories, alongside a collection of 600 actively reconstructed corpora. While this study used English data only, they emphasized the importance of considering multiple languages beyond one. 
Based on this dataset, a followup study \cite{sharma2023facilitating} developed a human language model interactive system, employing language models to assist individuals through cognitive reconstruction stages. 
While both studies underscore the detrimental effects of cognitive distortions, clarity on what kind of methods they used for distortion identification remains elusive. \par


\begin{figure*}[t]
  \centering
  \includegraphics[width=0.9\textwidth]{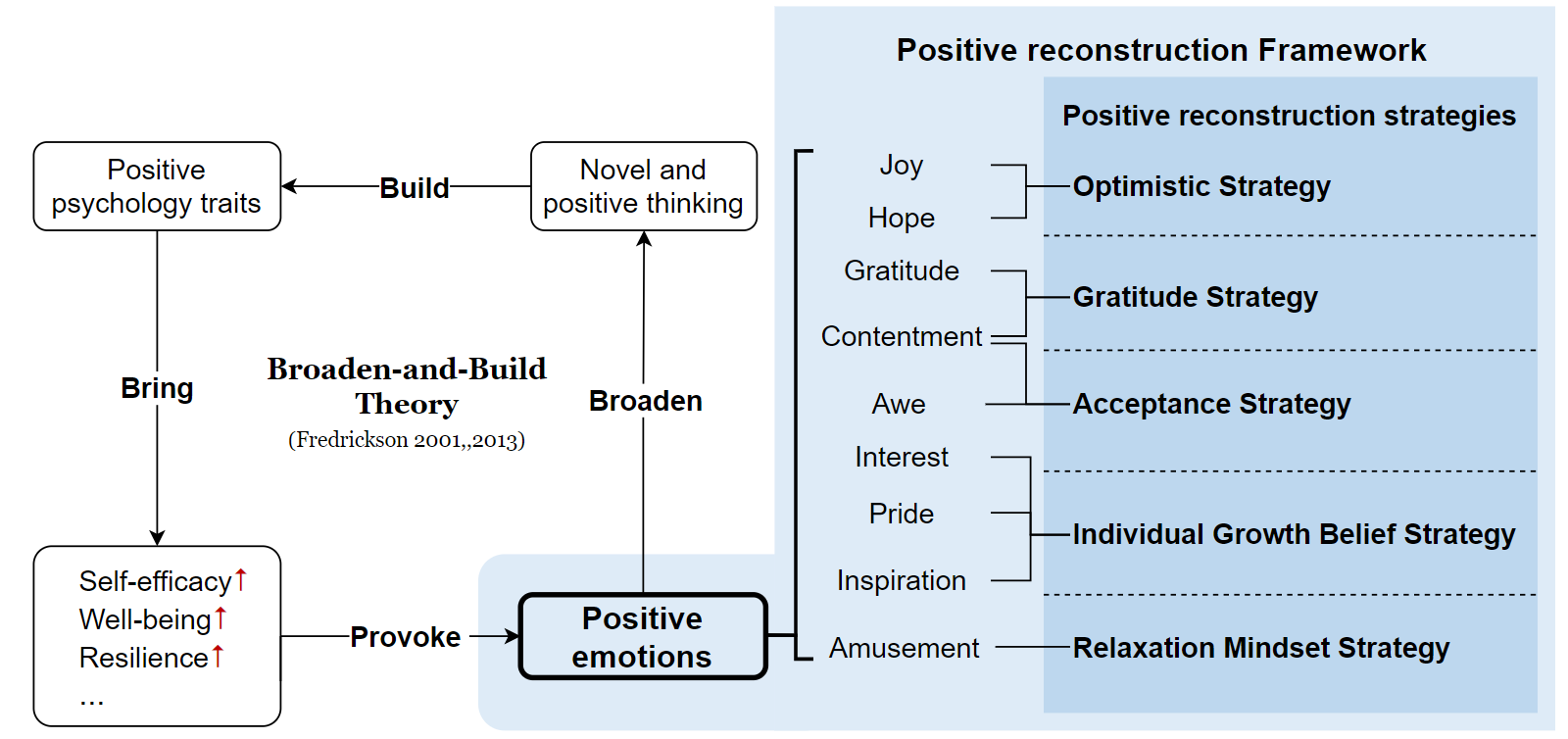}
  \caption{Broaden-and-build-theory \cite{fredrickson2001role,fredrickson2013positive} and how positive reconstruction can initiate this virtuous cycle}
  \label{fig:PRF}
\end{figure*}

\begin{table*}
  \centering
  \caption{Positive reconstruction strategies and associated positive emotions.}
  \begin{tabular}{l p{9cm} p{2.7cm}}
    \hline
    Strategy & Description & Emotions \\
    \hline
    Optimism &To be hopeful for the future while not overlooking or denying negative circumstances, shifting the focus towards more positive aspects. & joy, hope \\
    Gratitude & Developing a positive perspective on surrounding events and expressing gratitude. & thanksgiving, contentment \\
    Individual Growth & The belief that individuals can grow and have infinite potential within themselves.& interest, pride, inspiration \\
    Relaxation Mindset & Reducing the perceived harm of an event and relaxing tense negative emotions by understanding the limitations of negative experiences. & amusement \\
    Acceptance & Embracing one's imperfections and acknowledging others' abilities or efforts.& awe, contentment \\
    \hline
  \end{tabular}
  \label{tab:5ps}
\end{table*}

\section{Positive Reconstruction Framework}
\label{sec: Positive Reconstruction Framework}
This section presents five positive reconstruction strategies based on the broaden-and-build theory which we describe in this section. Utilizing principles derived from this theory can enhance the theoretical underpinnings of both present and forthcoming endeavours on the positive reframing task. These strategies attempt to reframe negative  thoughts from a positive perspective, fundamentally reorganizing rigid and harmful cognitive distortion beliefs.\par
The theory posits \cite{fredrickson2001role,fredrickson2013positive} that positive emotions broaden an individual's scope of attention, cognition, and action. Additionally, positive emotions can undo the effects of sustained negative emotions, known as the undoing hypothesis. They construct positive psychological traits, providing immediate and long-term benefits to individuals. 
Our study will propose positive reconstruction strategies based on the key positive emotions outlined in the theory, as shown in Table \ref{tab:5ps}.\par

\textbf{Optimistic Strategy}. According to the theory, joy refers to the feeling of happiness when unexpected good things happen in life and the anticipation of good things happening. Hope, on the other hand, represents an optimistic outlook towards the future. The difference between hope and joy lies in the fact that hope exists in stressful situations, where despite worries about the worst-case scenario, there is a simultaneous expectation for a better future. Therefore, we define the \textbf{optimistic strategy} as being hopeful for the future without disregarding nor denying the negative circumstances. 
For example, "I am so tired. Although I've been working so hard, I'll never succeed.,'' is a sentence that contains cognitive distortion, using optimistic strategy to reconstruct it can be: 
 "This process is truly exhausting, but I've put in a lot of effort and accumulated a lot. Continuing to persevere will yield positive results.''\par
\textbf{Gratitude Strategy}. In the theory, positive emotion gratitude refers to the feeling that arises when an individual receives something beneficial to themselves. Contentment, on the other hand, occurs when an individual cherishes their current circumstances. Both can be seen as perspectives of gratitude towards the present situation, aiding individuals in cultivating a positive outlook on surrounding events, known as the \textbf{gratitude strategy}. Using the previously distorted statement as an example, applying the gratitude strategy could rephrase it as, "Despite the hardship, I'm truly grateful for the opportunity to be part of this task. I've never experienced such high-intensity work before. Now I have more experience in dealing with pressure."\par
\textbf{Individual Growth Strategy} centers on the core belief that individuals can grow over time and have infinite potential within themselves. This links to key positive emotions in the theory such as interest, pride, and inspiration. 
Interest fosters an impulse for exploration and learning, allowing individuals to immerse themselves in novel experiences, expanding the self and facilitating personal growth. Pride often arises when individuals receive recognition, instilling a belief in their ability to achieve greater success in the future. Inspiration occurs when individuals witness others' commendable actions or outstanding performances, leading to a surge in self-motivation and a drive to surpass oneself, thus contributing to personal growth. These three positive emotions all contribute to fostering the belief that individuals can improve themselves. 
Employing this strategy to reconstruct statements with  distortions, such as "I procrastinate too much and never finish anything." can be transformed into "I can work on improving my time management skills." This shift in mindset has been proven to enhance students' academic performance \cite{yeager2014far,dweck2019mindsets}.\par
\textbf{Relaxation Mindset Strategy}. Understanding the limitations, impermanence, or commonality of errors reduces their perceived harm, thus easing tense negative emotions. For instance, when someone fails an exam, they might have thoughts like, "I failed the exam because I'm too stupid, I must be an idiot." However, the truth is that most people experience failures to some extent, and many have failed exams or even faced academic setbacks. According to the theory, amusement occurs when people encounter errors that pose small harm. Thus, a positive reconstruction to make the harmful error less severe could be, "I hope to perform better, but everyone makes mistakes; it's not just me who didn't pass."\par
\textbf{Acceptance Strategy}. According to the theory, contentment occurs when individuals feel comfortable within themselves or experience harmony with their external environment, it can be seen as self-acceptance. On the other hand, admiration arises when individuals encounter people or things that impress them, and a sense of awe motivates individuals to absorb and accept new experiences or individuals. Therefore, the \textbf{acceptance strategy} involves being able to calmly accept oneself without belittling others due to one's strengths or feeling inferior due to one's weaknesses.

\section{Dataset construction}
Previous research in cognitive distortion classification has revealed the challenges in achieving robust outcomes, necessitating ample data and high-quality annotation. Hence, we describe here the collection and annotation of a Chinese corpus with binary labels for the presence or absence of cognitive distortion, as well as positive reframing with five different strategies.  All collected and annotated data are shared online\footnote[1]{https://github.com/405200144/Dataset-of-Cognitive-Distortion-detection-and-Positive-Reconstruction/tree/main}.\par

\subsection{Data Collection}
The corpus labeling utilizes one specialized open-source dataset, the Chinese psychological Q\&A dataset PsyQA \cite{sun2021psyqa}, which is related to psychological issues and tends to contain statements with cognitive distortions.\par
This study aims to determine if the initial sentence input by the topic initiator is distorted, rather than the subsequent responses or discussions. The length of the sentences is restricted to between 10 and 66 characters to avoid excessively short or long and meaningless sentences. \par


\subsection{Annotation}
Two postgraduate students with a background in psychology annotated 4,001 pieces of corpus for cognitive distortion detection based on the Burns 10 classification standard \cite{burns1999feeling} (Appendix \ref{sec:appendix10CD}), 1129 of which were labeled as cognitive distortions, and internal consistency is 88\%.\par


The rewriting task was posted on the college student social platform. A total of 85 undergraduates and postgraduates participated in the pre-writing test. Finally, 42 participants passed the test and participated in the writing task, including 17 males and 25 females, from different majors, with an average age of 22.7. Each writer is requested to compose five distinct positive strategy reconstructions for one cognitive distortion sentence and complete 5 to 10 sets. In total, this process resulted in the collection of 1,900 sentences. 

\subsection{Data Quality}
In addition to using a pre-test to ensure that writers clearly understand the writing task (Appendix \ref{sec:appendixSS}), we also conducted a 5-class classification experiment, a sentiment assessment, and a manual evaluation to verify the quality of the writing. In the five-classification experiment, we split the training and test sets in a 4:1 ratio, using Chinese pre-trained model RoBERTa-wwm-ext \cite{cui2021pre} , an enhanced version of BERT \cite{kenton2019bert}, as the pre-trained model. The accuracy metric reached 82.85\% indicating that writers can compose different positive sentences based on various strategies. Furthermore, we utilized SnowNLP\footnote{https://github.com/isnowfy/snownlp} for sentiment positivity analysis of the corpus. The results indicated a notable uptrend for the manually written content. The average sentiment score of the original sentences was 0.72, while that of the manually written content was 0.96. In the manual evaluation, the average scores for handwritten submissions all exceeded 4 on a 5-point scale (Table \ref{tab:HER}). 


\section{Cognitive Distortion Detection}
In our study addressing the detection of cognitive distortions, we trained a binary classification model using a corpus of 4,001 entries annotated by individuals with a background in psychology. The sentences are fed into the neural network for a binary classification task to detect cognitive distortions. The dataset was randomly divided into a training set and a test set in a 4:1 ratio. Given the imbalanced nature of our dataset, where only approximately 28\% of the data represented distorted cognitions, we utilized focal loss to enhance the model's performance (with hyperparameters $\alpha = 0.25$ and $\gamma = 2$). We trained for 50 epochs with the AdamW optimizer and a learning rate equal to $10^{-5}$, determined using a coarse grid search. \par

We employ RoBERTa-wwm-ext as pre-trained model\cite{cui2021pre}. Two main strategies have been implemented to explore the performance of this language model for our particular task. First, transfer learning freezes all the weights in the pre-trained RoBERTa-wwm-ext network and augments it with a linear layer, a multilayer perceptron, or an LSTM model on the sequential RoBERTa-wwm-ext output. This last strategy is inspired by previous studies \cite{tan2022roberta,sirisha2022aspect} which have shown the potential of hybrid RoBERTa-LSTM for sentiment analysis. The multilayer-perceptron has 2 hidden layers of dimensions 512 and 256, with dropout rate at 0.2, and the LSTM hidden dimension is set to 256. \par

The second strategy is fine-tuning, in which we tune the weights of the whole RoBERTa-wwm-ext network with a linear classification readout for our classification task. This is more computationally-demanding and one has to be careful about catastrophic forgetting but it enables the network to learn better embeddings for the task at hand. \par

\begin{table}
\centering
\begin{tabular}{lcccc}
\hline
 &Acc&F1&R&P \\
\hline
Linear & 0.858 & 0.742 & 0.701 & 0.788 \\
    MLP & 0.863 & 0.775 & 0.808 & 0.744 \\
    LSTM & 0.874 & 0.781 & 0.769 & 0.793 \\
\hline
    Fine-tuning & \textbf{0.896} & \textbf{0.822} & \textbf{0.821} & \textbf{0.824} \\
\hline
\end{tabular}
\caption{\textbf{Results of Cognitive Distortion Detection} with linear, multilayer perceptron (MLP), LSTM readouts, and fine-tuning the pretrained RoBERTa-wwm-ext network, measured by Accuracy (Acc), F1 score (F1), Recall (R), and Precision (P)}
\label{tab:CD}
\end{table}

Results are presented in Table \ref{tab:CD}. We see that the best performing model is obtained with fine-tuning. The improvement is quite significant compared to transfer learning methods, which means that the embeddings by RoBERTa-wwm-ext are probably suboptimal for the cognitive distortion classification. Among the different transfer learning methods, using more complex readout layers improves the performance as we observe that the linear readout layer presents the worst performance. \par


\begin{table*}[h]
  \centering
  \caption{\textbf{Algorithmic evaluation results of positive reconstruction} measured by ROUGE-1, ROUGE-L, BLEU, BERTScore, and Sentiment analysis with SnowNLP, for P-tuning (V2) of ChatGLM-6B, Fine-tuning of ChatGPT3.5, and prompt engineering of both models.} 
  \begin{tabular}{cc|ccccc}
    \hline
  
    \multirow{2}{*}{\textbf{Method}}&\multirow{2}{*}{\textbf{Model}}&\multicolumn{5}{c}{\textbf{Auto evaluation}} \\
    &&ROUGE-1&ROUGE-L&BLEU&BERTScore&Sentiment\\
    \hline
  \multirow{2}{*}{\textbf{Fine-tune}}&ChatGLM-6B&\textbf{31.89}&\textbf{27.14}&10.67&\textbf{32.25}&0.91 \\
    
 &ChatGPT3.5 &30.00&25.70&\textbf{10.68}&29.94&\textbf{0.98} \\
  \hline
 \multirow{2}{*}{\textbf{Prompt}}&ChatGLM-6B&24.69&19.04&6.70&23.51&0.93 \\

&ChatGPT3.5&28.73&23.69&9.98&28.09&0.93 \\

    \hline
  \end{tabular}
  \label{tab:AER}
\end{table*}

\section{Positive Reconstruction}
The emergence of large-scale language models has revolutionized various tasks, enabling their completion through prompt engineering \cite{mesko2023prompt, liu2022design}. This led us to wonder whether the performance of prompt engineering parallels or surpasses that of fine-tuning for positive reconstruction generation. 


\subsection{Experimental Setup}

Compared to the binary classification for cognitive distortion detection, the task at hand involves conversational Large Language Models for text generation. We leverage the recent developments with open-source models and APIs to benchmark 4 different approaches:
\begin{itemize}
    \item P-tuning (V2) of ChatGLM-6B \cite{du2022glm}
    \item Fine-tuning of ChatGPT-3.5 Turbo-1106 \cite{ouyang2022training}
    \item Prompt engineering with ChatGLM-6B 
    \item Prompt engineering with ChatGPT-3.5
\end{itemize}

\subsubsection{Prompt engineering}
We provided each model with one set of examples shown in Appendix \ref{sec:appendixPrompt}. 
We installed ChatGLM-6B by following their instructions on Github\footnote{https://github.com/THUDM/ChatGLM-6B}, and accessed the API\footnote{https://platform.openai.com/docs/api-reference} from the free version of ChatGPT. Prompt engineering offers the advantage that it can be used with closed models, even in situations where access is limited to model outputs rather than weights. Moreover, as it is based on generalist language models, it does not require a parallel (or supervised) dataset, only the current sentence is required. 



\subsubsection{Training ChatGLM-6B Based on P-Tuning (V2)}

P-Tuning v2 \cite{liu-etal-2022-p} is based on Deep Prompt Tuning \cite{lester2021power,qin2021learning} designed for generation and knowledge probing. For a large language model with 300M to 10B parameters, P-Tuning v2 can achieve similar results as Fine Tuning by introducing trainable "prompts" to each layer in the language model, but it only requires the tuning of 0.1\% to 3\% of the parameters, which greatly reduces the cost of training. 

ChatGLM-6B is trained with a batch size of 1 and global training steps of 3,000, and a learning rate of 0.02 is applied. 
We randomly split the datasets in an 8:1:1 ratio for training, validation, and testing. The maximum input and target token lengths are set to 64, tailored for concise and focused input-output relationships. Additionally, 4-bit quantization is employed, optimizing the model for efficient computation. Optimization on an NVIDIA RTX A4000 GPU (16 GB) took approximately 5 hours. 

\begin{table}[h]
  \centering
  \caption{\textbf{Human evaluation results} in the preservation of meaning (M), positivity (P), and an overall appreciation (O) of the reconstructed sentences. Comparison between P-Tuning (V2) of ChatGLM-6B (GLM), Fine-tuning of ChatGPT3.5 (GPT3.5), prompt engineering of both models, and ground truth.}
  \begin{tabular}{cc|ccc}
    \hline
  
    \multirow{2}{*}{\textbf{Method}}&\multirow{2}{*}{\textbf{Model}}&\multicolumn{3}{c}{\textbf{Human evaluation}} \\
    &&M&P&O\\
    \hline
  \multirow{2}{*}{\textbf{Fine-tune}}&GLM&\textbf{4.08}&\textbf{3.88}&\textbf{3.87} \\
    
 &GPT3.5 &3.05&3.25&2.95 \\
  \hline
 \multirow{2}{*}{\textbf{Prompt}}&GLM&2.92&3.29&2.87 \\

&GPT3.5&3.57&3.79&3.60 \\
\hline
\multicolumn{2}{c|}{Ground Truth}&4.04&4.17&4.02 \\
   
    \hline
  \end{tabular}
  \label{tab:HER}
\end{table}

\subsubsection{Training ChatGPT 3.5 Turbo-1106 Based on Fine-tuning}


For the fine-tuning process of ChatGPT 3.5 Turbo-1106, we harnessed OpenAI's existing API. According to OpenAI\footnote{https://platform.openai.com/docs/guides/fine-tuning (Accessed December 2023)}, their fine-tuning API enhances few-shot learning by training on a significantly larger set of examples than can be accommodated within the prompt. This approach enables users to achieve improved results across a wide array of tasks. Once a model has been fine-tuned, users will not need to provide as many examples in the prompt. 

\subsection{Evaluation}
Our research involved assessing the performance of our models in terms of their semantic similarity to the original text. Following other sentence reconstruction work \cite{ziems2022inducing,sharma2023cognitive}, we used several metrics for this evaluation: BLEU \cite{papineni2002bleu}, ROUGE-1, ROUGE-L \cite{lin2004rouge}, and BERTScore \cite{zhang2019bertscore}. Given that each cognitive distortion sentence had five reconstructed ground truth annotations, we documented these independently and selected the highest results to put in the Table \ref{tab:AER}. Additionally, we determined the sentiment score using SnowNLP.\par
We conducted a human evaluation to obtain more insights about the ground truth and model-generated sentences. Each participant was asked to rate 50 sentences on a five-point Likert scale based on three aspects: the preservation of meaning, the positivity, and the overall appreciation of the sentences. A total of 29 evaluation feedbacks from non-research-related personnel were received.\par


\begin{table}[ht]
  \centering
  \caption{\textbf{Algorithmic evaluation results} of sentences generated by ChatGLM-6B P-Tuning (V2) compared with the ground truth of five different strategies.}
  \begin{tabular}{lcccc}
    \hline
    
Strategy& R-1&R-L&BLEU&BScore \\
    \hline
   
     Optimism & \textbf{31.89}&\textbf{27.14}&\textbf{10.67}&\textbf{32.25} \\
     Gratitude  & 29.82 &25.16&10.11&30.56\\
     Growth & 31.44 &26.40&10.35&31.73\\
     Relaxation& 30.46 &25.29&9.97&29.36\\
     Acceptance & 29.69 &24.31&9.05&30.24\\

    \hline
  \end{tabular}
  \label{tab:5S}
\end{table}

\subsection{Results and Analysis}
In the automatic sentiment evaluation results (Table \ref{tab:AER}), it was found that prompt results generated by ChatGPT-3.5 (GPT-3.5) achieved the highest score of 0.98. However, the difference between the four models was minimal, and all showed a significant improvement in positivity compared to the negative original sentences 0.72. On the other hand, in the aspect of semantic similarity, ChatGLM-6B (GLM) P-Tuning scored the best results across all four metrics, followed by GPT-3.5 fine-tuning, with GLM prompt performing the worst. These results indicate that fine-tuning large language models performs better than prompt engineering for positive reconstruction tasks. As mentioned earlier, this underscores the significance of possessing a parallel dataset for the process of supervised fine-tuning. Interestingly, fine-tuning yields better results than prompt engineering, a finding that diverges from the conclusion presented in \cite{maddela2023training}. This underscores the importance of tailoring approaches to individual tasks.\par
Additionally, we also observed that GLM P-Tuning outperformed GPT-3.5 fine-tuning, a finding that aligns with a previous study \cite{hu2023llm}, which found that in certain specialized tasks, smaller language models, such as LLaMA-13B and LLaMA-7B, demonstrated superior performance compared to larger counterparts like ChatGPT. At the same time, it is important to point out here that GPT-3.5 Fine-Tuning results show limited improvement over Prompt Engineering. This aligns with the observation in \cite{lester2021power} that fine-tuning is relatively effective only for models with less than 10 billion parameters.\par

We present in Table \ref{tab:5S} an algorithmic evaluation of the sentences generated by the best performing model, ChatGLM-6B with P-Tuning (v2), to compare them with the different strategies employed. This makes use of our dataset which also classifies each positive reconstruction according to its reframing strategy. We also observed that the outputs of each model were more similar to the optimistic ground truth as shown in Table \ref{tab:5S}. Given that the proportions of the five strategies in the fine-tune training set were similar, this result may suggest that large language models are inherently more likely to generate statements similar to the optimistic strategy, or that the training corpus with different strategies contains more sentences that resemble the optimistic strategy. This call for further exploration and research to analyze the bias in the outputs of language models. \par
The results of the human evaluation (Table \ref{tab:HER}) were similar to the automatic assessment. Among the four models, GLM P-Tuning performed the best, even exceeding the manually written scores in terms of preserving the original meaning, while GLM prompt was the worst. For GPT-3.5, automatic evaluation metrics are higher with the fine-tuning strategy, while the prompt engineering version performs better for human evaluation. 
The reason for this discrepancy is also something we hope to explore further.

\section{Discussion}
In conclusion, this study demonstrated the efficacy of NLP techniques in detecting and reconstructing cognitive distortions. We introduced a Positive Reconstruction Framework, grounded in positive psychology theory, and developed a dataset based on its five strategies. This framework was applied to fine-tune large language models, and its performance was compared with that of prompt engineering. Our research indicates that fine-tuning these models is more effective than prompt engineering.\par
In future work, a key focus will be conducting extensive user testing experiments to determine whether sentences reconstructed with a positive approach can effectively alleviate negative emotions or stimulate positive ones in users. Additionally, exploring the vast potential of prompt engineering to identify the most effective prompts for optimal resource utilization will be a priority. Also, further research into how the size of large language models influences their fine-tuning performance in specific domain tasks will be crucial, offering a comprehensive understanding of their application in practical scenarios.

\section*{Ethics}
The data used for this study was obtained from an online public psychological Q\&A, which contains some overly negative sentences that may have a negative impact on annotators. To mitigate this potential risk, we made sure that all contributors were 18 years or older and none of them would be affected by any negative statements before starting the annotation task. Those with mental health issues, who felt they might be susceptible to such issues, or who could be influenced by negative emotions severely, were not permitted to participate in the data construction tasks.

\section*{Limitations}

While this study tackles the problem of language-specific datasets and algorithms, it should be noted that our focus was primarily on the Chinese language. Future research is needed to include a more diverse range of languages, thereby providing a more comprehensive understanding of cognitive restructuring across different linguistic and cultural contexts.

Furthermore, it's important to acknowledge that this work is still far from any immediate application. While we have made strides in constructing the Mandarin dataset and benchmarking various NLP models, future studies with more refined human evaluation, investigating biases and risks, are required before it can directly benefit individuals seeking mental health support. 

Additionally, it's worth noting that some of the NLP models utilized in this study are only accessible through paid APIs, limiting their accessibility to researchers and practitioners without the necessary resources. 

\section*{Acknowledgments}

We thank the anonymous reviewers for their insightful feedback and suggestions. This work is based on work supported by the Chinese National Social Sciences Foundation (No. BBA210042).

\bibliography{custom}

\clearpage
\appendix
\onecolumn

\section{Positive Reconstruction}
\label{sec:appendixPRE}

\begin{table}[h]
  \centering
  \caption{\textbf{A model comparison example of positive reconstruction} among the same original text measured by ROUGE-1(R-1), ROUGE-L(R-L), BLEU, BERTScore, Sentiment analysis, and human evaluation with the different models, along with English translations and associated scores. GLM-6B and GPT3.5 represent ChatGLM-6B and ChatGPT3.5, respectively.}
  \begin{tabular}{c|cccccccc}
    \hline
    \multicolumn{9}{l}{\textit{Original Text: I'm not sure what to do either;}} \\
    \multicolumn{9}{r}{\textit{ I find myself getting angry easily every day and taking it out on others.}} \\
    \multicolumn{1}{c}{}&\multicolumn{8}{c}{\begin{minipage}[b]{0.8\columnwidth}
		\centering
		\raisebox{-.1\height}{\includegraphics[width=\linewidth]{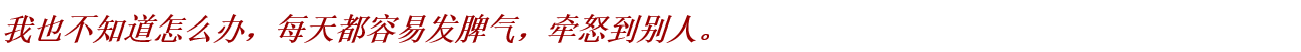}}
	\end{minipage}}
 \\
    \hline
    \multirow{2}{*}{Model}&\multicolumn{5}{c}{\textbf{Algorithmic evaluation}} &\multicolumn{3}{|c}{\textbf{Human evaluation}} \\
    &R-1&R-L&BLEU&BERTScore&Sentiment&\multicolumn{1}{|c}{Meaning}&Positivity&Overall\\
    \hline
  \multirow{4}{*}{\shortstack{GLM-6B \\P-Tuning (V2)}}&\textbf{41.43}&\textbf{34.81}&6.91&\textbf{43.88}&0.93&\textbf{4.38}&3.46&\textbf{3.85} \\
     &\multicolumn{8}{c}{\multirow{2}{*}{\textit{\shortstack{Although I easily lose my temper and take my anger out on others every day,\\ I believe I can try to change and become a person with stable motions.}}}} \\
    &\multicolumn{8}{c}{} \\
    &\multicolumn{8}{c}{\begin{minipage}[b]{0.8\columnwidth}
		\centering
		\raisebox{-.1\height}{\includegraphics[width=\linewidth]{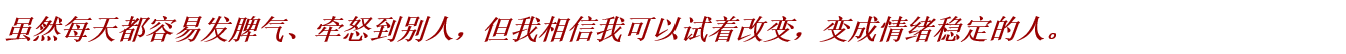}}
	\end{minipage}}
 \\
        \hline
 \multirow{5}{*}{\shortstack{GPT3.5 \\Fine-Tuning}}&34.31&24.97&7.37&35.87&0.99&3.69&3.46&3.46 \\
     &\multicolumn{8}{c}{\multirow{2}{*}{\textit{\shortstack{Although I easily lose my temper every day, everyone develops from immaturity,\\ and I can continue to improve my emotional control and communication skills.}}}} \\
    &\multicolumn{8}{c}{} \\        
   &\multicolumn{8}{c}{\begin{minipage}[b]{0.8\columnwidth}
		\centering
		\raisebox{-.1\height}{\includegraphics[width=\linewidth]{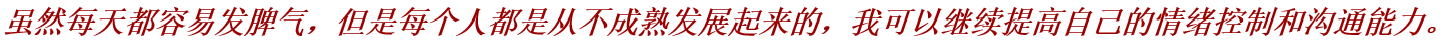}}
	\end{minipage}}
 \\
        \hline
  \multirow{7}{*}{\shortstack{GLM-6B\\Prompt}}&24.66&16.41&7.16&28.79&\textbf{1.00}&2.85&\textbf{3.77}&3.00 \\
      &\multicolumn{8}{c}{\multirow{5}{*}{\textit{\shortstack{Learn to control your emotions and try to reduce the frequency\\ of losing your temper. If you find it difficult to control your emotions, \\consider techniques like deep breathing or meditation to relax.\\ You could also try talking to a close friend or family member,\\ sharing your feelings and emotions to seek their advice and support.}}}} \\
    &\multicolumn{8}{c}{} \\
    &\multicolumn{8}{c}{} \\
    &\multicolumn{8}{c}{} \\
    &\multicolumn{8}{c}{} \\
  &\multicolumn{8}{c}{\begin{minipage}[b]{0.8\columnwidth}
		\centering
		\raisebox{-.1\height}{\includegraphics[width=\linewidth]{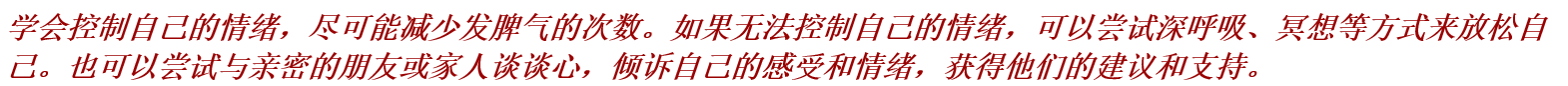}}
	\end{minipage}}
 \\
        \hline
    \multirow{5}{*}{\shortstack{GPT3.5\\Prompt}}&38.97&33.15&\textbf{10.44}&43.44&0.99&3.54&3.54&3.80 \\
     &\multicolumn{8}{c}{\multirow{4}{*}{\textit{\shortstack{I’ve realized that I’ve been quick to anger lately. I will try to \\alleviate my emotions by learning relaxation techniques, such as\\ practicing yoga or meditation. I am committed to managing\\ my emotions better and avoiding taking my anger out on others.}}}} \\
    &\multicolumn{8}{c}{} \\
    &\multicolumn{8}{c}{} \\    
    &\multicolumn{8}{c}{} \\
 &\multicolumn{8}{c}{\begin{minipage}[b]{0.8\columnwidth}
		\centering
		\raisebox{-.1\height}{\includegraphics[width=\linewidth]{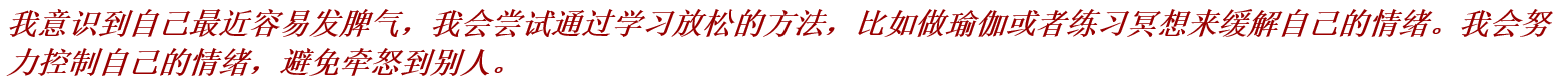}}
	\end{minipage}}
 \\
        \hline
    
  \end{tabular}
  \label{tab:PRE}
\end{table}

\clearpage
\section{Cognitive Distortion Detection Annotation}
\label{sec:appendix10CD}
This task was conducted by two postgraduate students with a background in psychology. The goal was to distinguish whether texts contain any distortion according to 10 Cognitive Distortion type of \citet{burns1999feeling}, annotators gave the label 0 as non-distortion, and 1 as distortion.

\begin{table}[h]
  \centering
  \caption{10 Cognitive Distortion type of \citet{burns1999feeling}}
  \begin{tabular}{ll}
    \hline
    Distortion type&Description \\
    \hline
    Emotional Reasoning & Reasoning base on emotions, believing “I feel that
way, so it must be true''\\
    Overgeneralization & Drawing conclusions with limited and often un negative experience.\\
    Mental Filter & Focusing only on limited negative aspects and not the excessive positive ones.\\
    Should Statement & Expecting things or personal behavior should be a certain way.\\
    All or Nothing& Binary thought pattern. Considering anything short of perfection as a failure.\\
    Mind Reading& Concluding that others are reacting negatively to you, without any basis in fact.\\
    Fortune Telling& Predicting that an event will always result in the worst possible outcome.\\
    Magnification& Exaggerating or Catastrophizing the outcome of certain events or behavior.\\
    Personalization& Holding oneself personally responsible for events beyond one’s control.\\
   Labeling & Attaching labels to oneself or others (ex: “loser”, “perfect”).\\
    \hline
  \end{tabular}
  \label{tab:10CD}
\end{table}

\section{Data Collection Interface Snapshots and English translation}
Before conducting the positive rewrite annotation task we set a pre-test to ensure that annotators can understand the task correctly.
\label{sec:appendixSS}

\begin{figure}[h]
  \centering
  \includegraphics[width=1\linewidth]{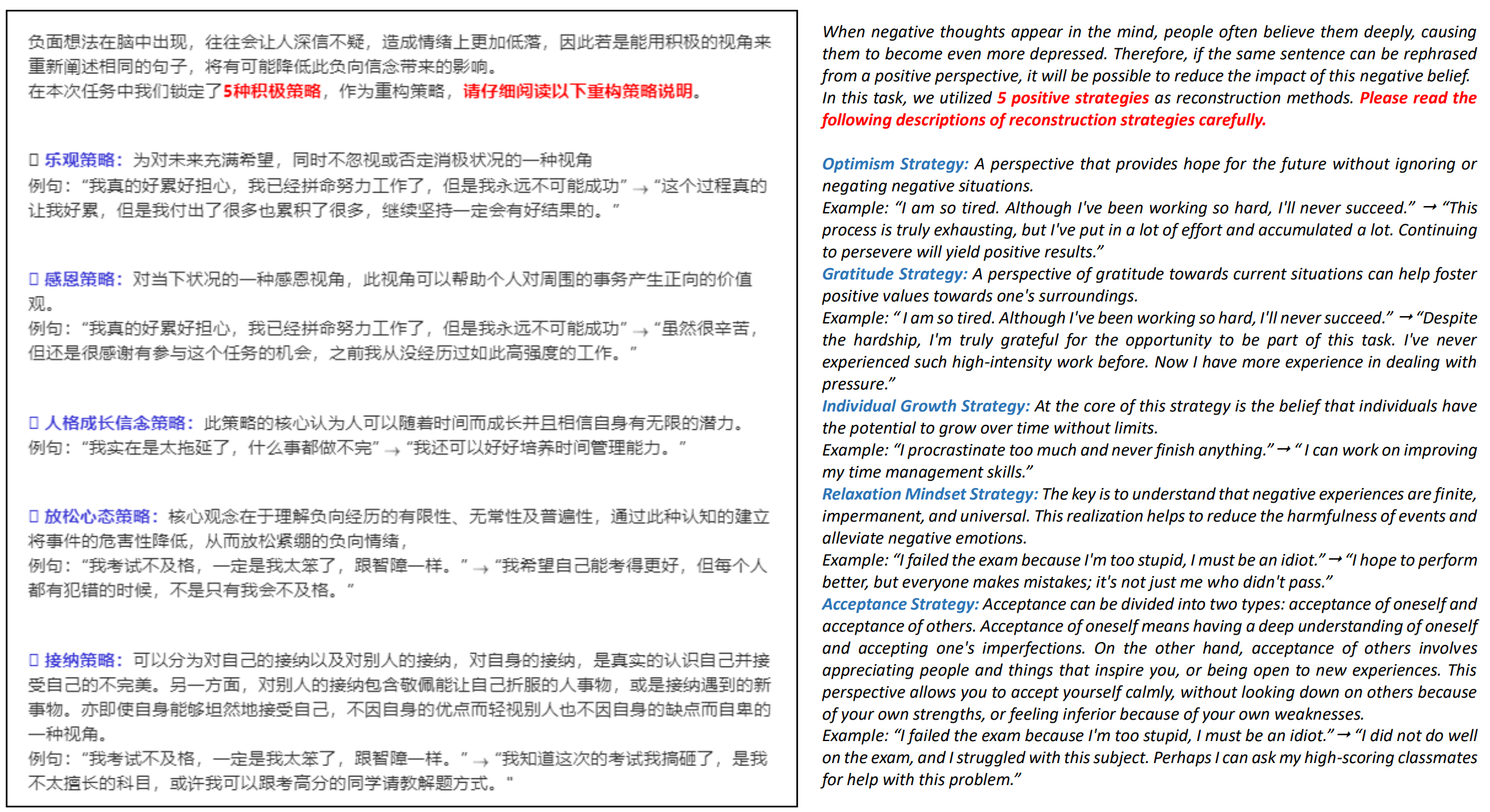}
  \caption{Description of the 5 positive reconstruction strategies, this was shown in pre-test and the official rewriting task.}
  \label{fig:f5.2}
\end{figure}
Two types of questions were presented in the pre-test. The first part consisted of 7 questions. We asked participants to choose the strategies they thought were most likely to be used by providing one negative and one positive sentence. Followed by answering three questions on choosing the most qualified rewritten sentence from four options in the second part. The goal was to ensure that the revised sentence conveyed the original meaning while avoiding overgeneralization.
Participants who answered correctly to more than 8 questions can participate in the writing task.

\begin{figure}[h]
  \centering
  \includegraphics[width=1\linewidth]{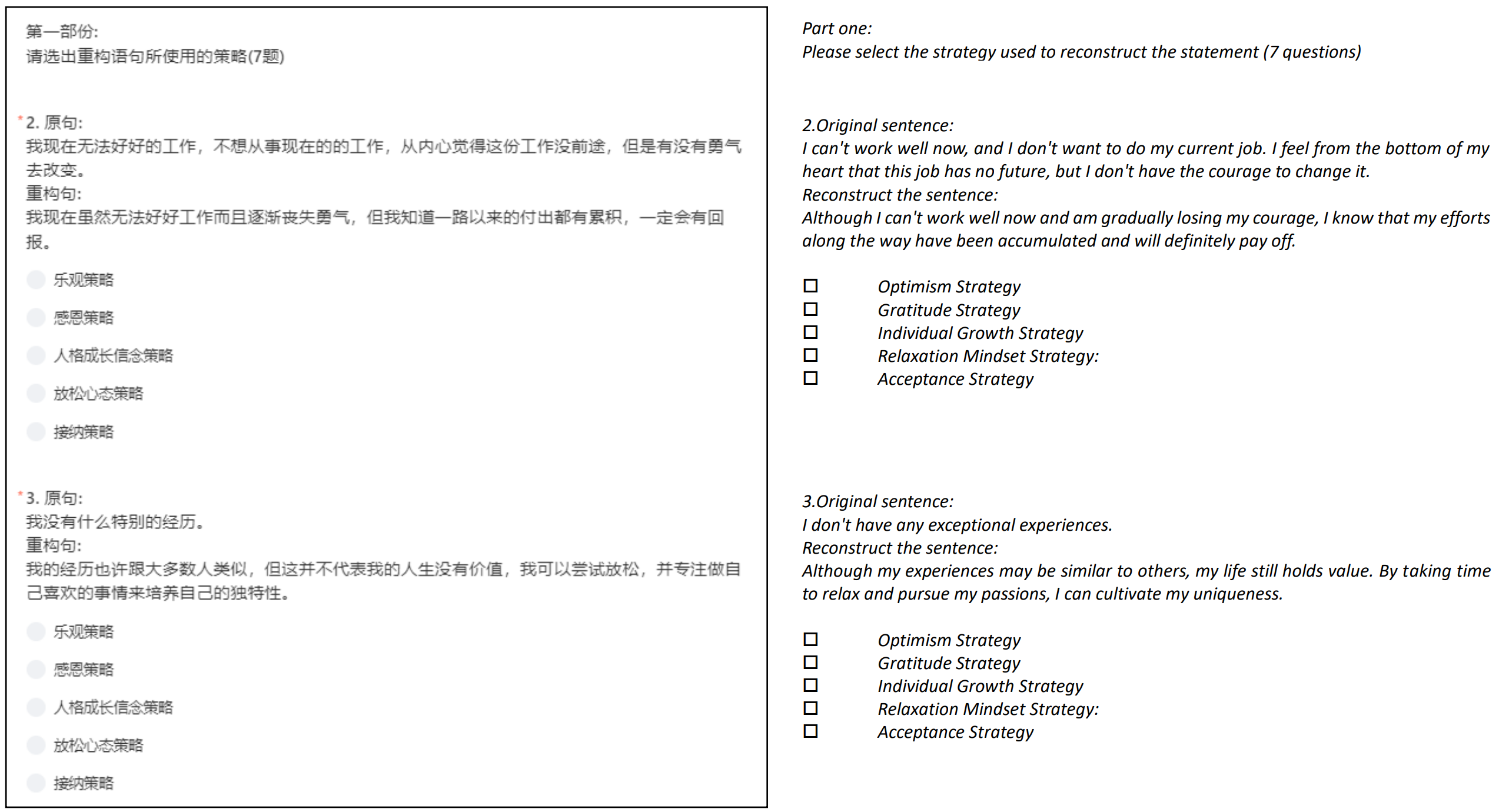}
  \caption{Example of distinguishing different strategies in pre-test.}
  \label{fig:f5.3}
\end{figure}

\begin{figure}[h]
  \centering
  \includegraphics[width=1\linewidth]{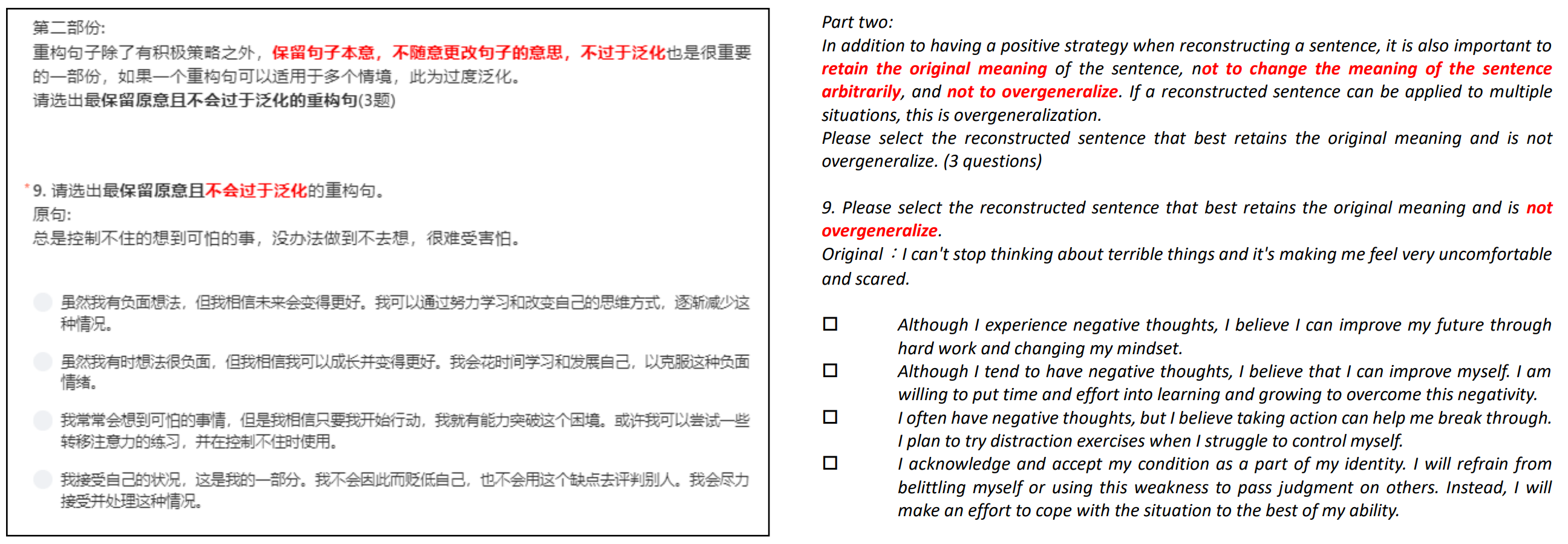}
  \caption{Example of selecting the most qualified rewritten sentence in pre-test.}
  \label{fig:f5.4}
\end{figure}

\begin{figure}[h]
  \centering
  \includegraphics[width=1\linewidth]{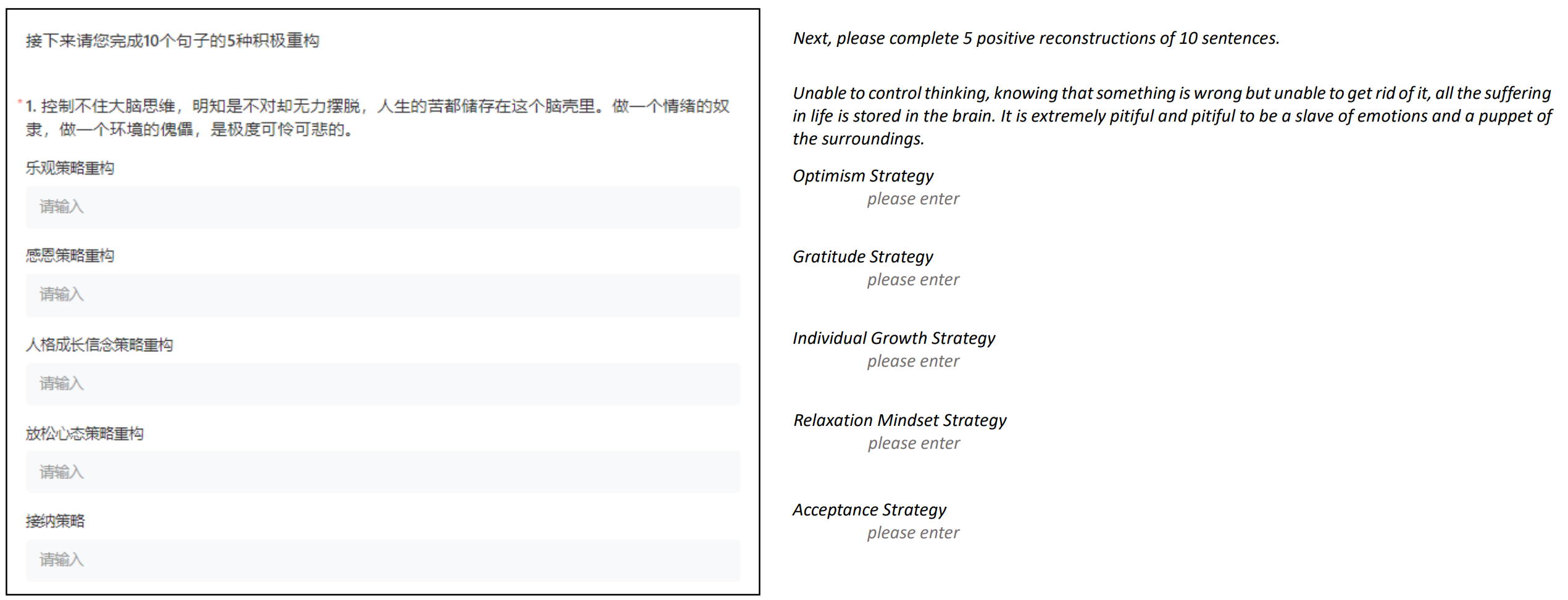}
  \caption{Example of positive rewriting task.}
  \label{fig:f5.5}
\end{figure}

\clearpage

\begin{figure}[ht]
  \includegraphics[width=1\linewidth]{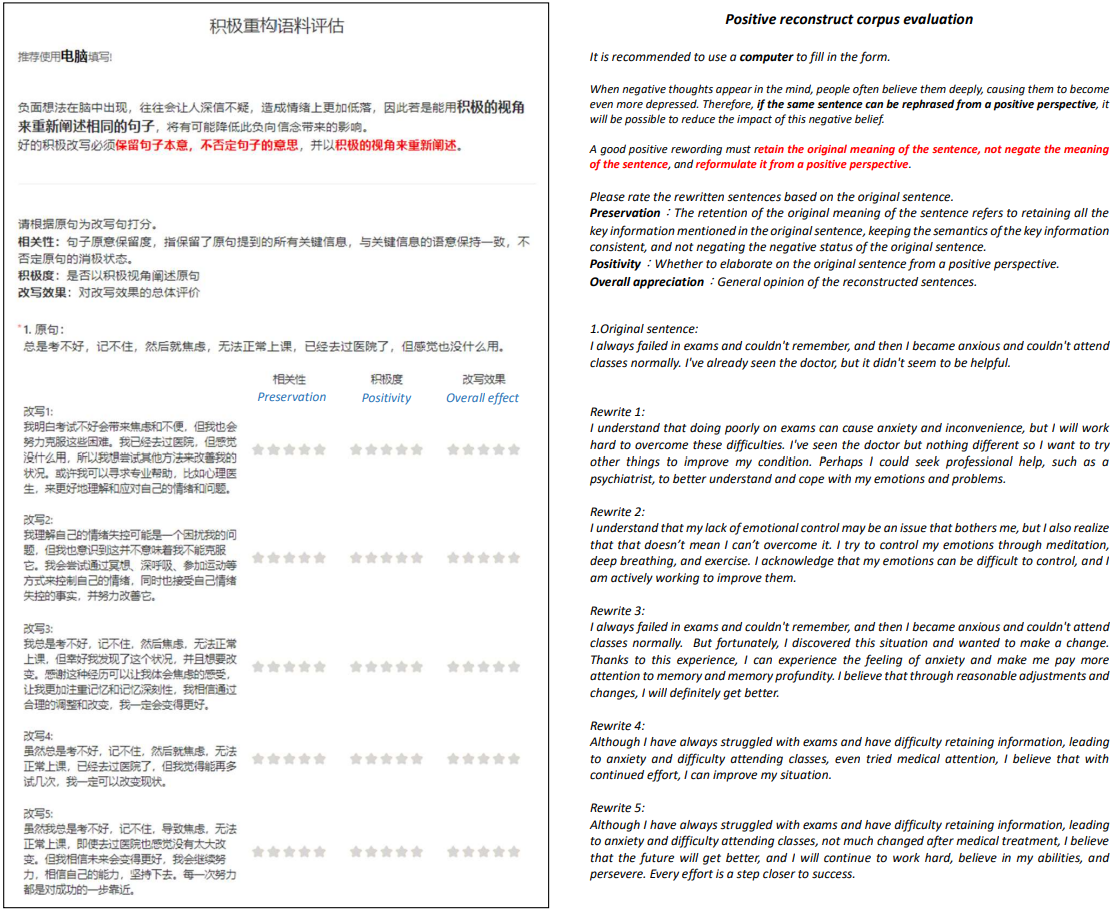}
  \caption{Example of Human Evaluation Interface Snapshots and English translation.}
  \label{fig:f5.6}
\end{figure}

We used a free questionnaire platform\footnote{https://www.wenjuan.com/} to set up pre-tests, rewrite tasks and evaluate questionnaires.

\clearpage

\section{Example Prompt for the Task of Reframing Negative Thoughts}
\label{sec:appendixPrompt}

\begin{figure}[h]
  \centering
  \includegraphics[width=0.65\textwidth]{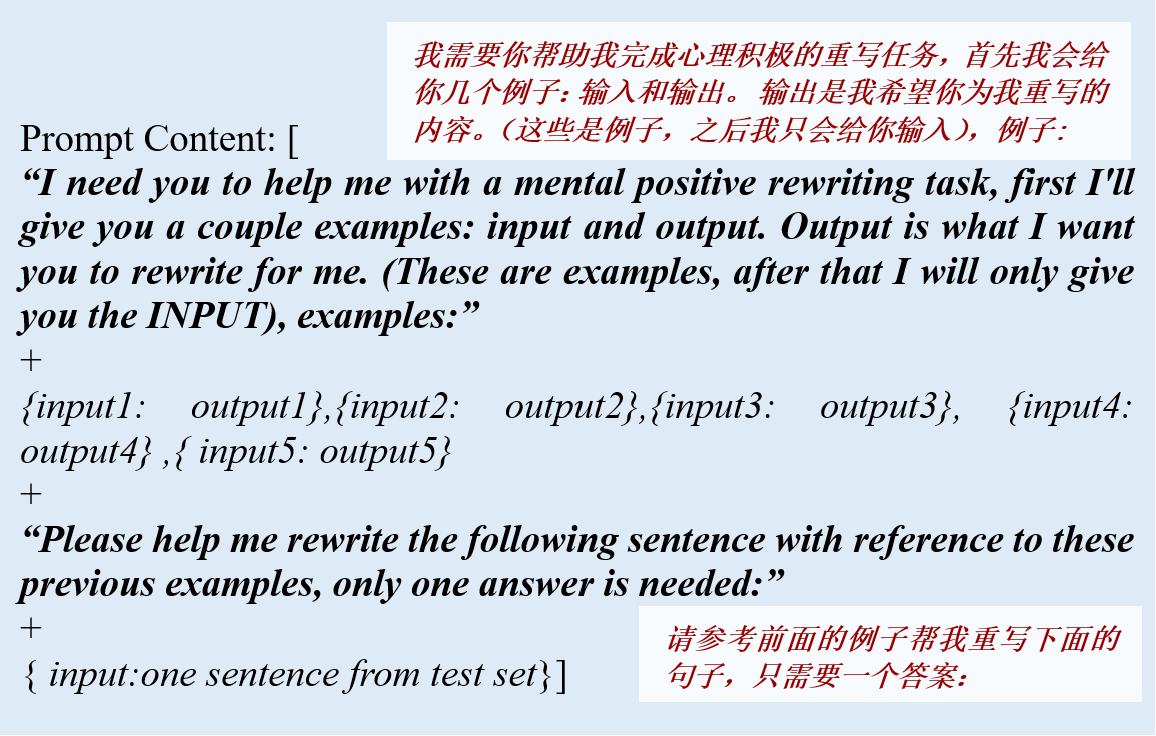}
  \caption{The translated English prompt example used for text reconstruction based on single turn conversation: (same for both ChatGPT3.5 and ChatGLM-6B). Specifically, we give five examples based on the five strategies, and finally we give the intended reconstruction sentence at the end of the prompt. We implement prompt engineering using the test dataset.}
  \label{fig:prompt}
\end{figure}




\clearpage

\end{document}